\documentclass[runningheads]{llncs}

\usepackage{graphicx}
\usepackage[utf8]{inputenc}
\usepackage{amsmath}
\usepackage{url}
\usepackage[linesnumbered,ruled,vlined]{algorithm2e}
\usepackage{amsfonts}
\usepackage{amssymb}
\usepackage{tabularx}
\usepackage{nicefrac}
\usepackage{stmaryrd}
\usepackage{bbm}
\begin{document}

\title{Cold Start Active Learning Strategies in the Context of Imbalanced Classification}
\titlerunning{Cold Start Active Learning}

\institute{}
\author{}

\institute{Luxembourg Institute of Science and Technology, Luxembourg \\
\email{firstname.lastname@list.lu} \and
University of Geneva, Switzerland \\
\email{firstname.lastname@unige.ch}}

\author{Etienne Brangbour\inst{1,2} \and
Pierrick Bruneau\inst{1} \and
Thomas Tamisier\inst{1} \and \\
Stéphane Marchand-Maillet\inst{2}}
\authorrunning{E. Brangbour et al.}

\institute{Luxembourg Institute of Science and Technology, Luxembourg \\
\email{firstname.lastname@list.lu} \and
University of Geneva, Switzerland \\
\email{firstname.lastname@unige.ch}}

\maketitle
\begin{abstract}
We present novel active learning strategies dedicated to providing a solution to the cold start stage, i.e. initializing the classification of a large set of data with no attached labels. Moreover, proposed strategies are designed to handle an imbalanced context in which random selection is highly inefficient. Specifically, our active learning iterations address label scarcity and imbalance using element scores, combining information extracted from a clustering structure to a label propagation model.
The strategy is illustrated by a case study on annotating Twitter content w.r.t. testimonies of a real flood event. We show that our method effectively copes with class imbalance, by boosting the recall of samples from the minority class.

\keywords{Active Learning \and Cold Start \and Imbalanced Data \and Label Propagation.}
\end{abstract}

\section{Introduction}

Learning a classification model usually happens by fitting a model (e.g. multilayer neural networks \cite{glorot_understanding_2010}, SVM \cite{scholkopf_learning_2001}) using independent labeled training and validation data sets. Trained models are then used in a production environment, where labels are unknown and to be predicted.
However, in real applications, while collecting and storing large amounts of unlabeled data has become fairly straightforward, they come without supervision. The labels are often implicit, i.e. only experts or users are able to tell to which class a given data item belongs to. Also, this operation is time consuming and costly.

Methods from the active learning domain are aimed at supporting such situations \cite{settles_active_2009}. In brief, starting from an initial classification model estimated using a small subset of labeled data (comparatively to the amount of unlabeled data that has been collected), the core of active learning is to design strategies for sampling data items from the large unlabeled collection in a way that will be most likely to improve the classifier. Such data are submitted to an oracle for annotation, per batches or one by one. Returned labeled elements are integrated to the model training set and the classification model is retrained. The procedure is iterated until some convergence criterion is matched, or a predefined budget for annotation is consumed.

Imbalanced training sets are problematic for learning classification models, and are addressed with specific solutions \cite{he_learning_2009}. However, in the context of active learning, especially at the cold start stage where no labels are available, this yields a \emph{chicken and egg problem}: sampling an initial training set at random is highly likely to yield only negative elements. The initial resulting classifier will hence be poor, harming any active learning strategy. This leads to poor convergence, that contradicts the spirit of active learning that aims at \emph{efficient labeling}. This is critical especially in a crowdsourcing context, where annotations are tied to a financial cost, and labeling many redundant elements should be avoided.

As an answer to this typical \emph{needle in the haystack} problem, we propose novel active learning strategies and algorithms that jointly exploit a clustering structure and a semi-supervised label propagation model, as means to increase chances to label minority class elements early in the process, while sampling from the overall data distribution in a faithful way. After recalling the related work, we expose the rationale of our methodology in Section \ref{sec:methodology}. Specifically, after emphasizing elements from the literature that serve as our cluster cluster quality and label propagation step, we introduce our contributed algorithms and strategies, that build upon these elements. A real use case, along with data collection and preprocessing guidelines, are then exposed in Section \ref{sec:experiments}. The use case serves as experimental testbed to evaluate the effectiveness of our approach, and the relative performance of proposed algorithmic variants.

\section{Related Work}

Imbalance occurs when severe class distribution skew is observed in the labeled data, or some classes are under-represented \cite{he_learning_2009}. Without adaptation, fitted models tend to reach 100\% accuracy for the majority class, but close to 0\% for the minority class. Most methods are based on oversampling the minority class or undersampling the majority class. Informed undersampling is based on model ensembling or kNN classification. For over-sampling, instead of basic replacement, synthesis according to nearest neighbor information may be considered \cite{chawla_smote:_2002}, or be combined with boosting \cite{chawla_smoteboost_2003}. The authors define data set complexity as its tendency to feature overlapping classes and multiple subconcepts. In the case of imbalanced data, relevant metrics are precision and recall, ROC curves, and cost curves.

Closely related to cold start situations, few-shot learning is presented as the context where a limited number of labeled examples is available \cite{wang_generalizing_2020}. A possible scenario is the reduction of labeled data gathering effort. In their taxonomy, active learning and semi-supervised learning are known as two instances of weakly supervised learning. To circumvent the difficulty to learn with limited supervision, methods typically resort to prior knowledge. Semi-supervised learning and active learning are based on the exploitation of unlabeled data.
Douze et al. apply a semi-supervised label propagation algorithm \cite{zhu_semi-supervised_2003} for few-shot learning with large image collections \cite{douze_low-shot_2018}. Few labeled and many unlabeled images are embedded using a pre-trained convolutional model. A semi-supervised model then extends the labeling to the unlabeled data. They use large similarity graphs based on nearest neighbors for propagation, with the claimed possibility to scale up to millions of elements using an adapted algorithm and library \cite{johnson_billion-scale_2019}.

When semi-supervised approaches aim at getting the most out of a small labeled set, active learning considers a dynamic setup, where labels are acquired in sequence \cite{settles_active_2009}.
Active learning iterations embody this setup, with batches of elements queried for labels to an oracle. The active learning strategy defines how these elements are chosen at each iteration.
For instance, uncertainty sampling strategies query elements in regions where a tentative classifier, trained with the currently available labeled data, is the least confident \cite{lewis_sequential_1994}. Query-by-committee uses ensembles of models to reduce disagreement among models \cite{seung_query_1992}.
The semi-supervised approach by Wu et al. \cite{wu_exploit_2018} exploits the most confident predictions, and differs significantly from active learning schemes such as uncertainty sampling, where elements are selected for query in the most uncertain areas of the data space. This highlights that semi-supervised approaches cannot be adapted to active learning contexts in a straightforward way.

Ertekin et al. implement active learning in the context of a SVM \cite{ertekin_learning_2007}. They implement a strategy inspired by uncertainty sampling, by querying elements in the vicinity of the separating hyperplane characterizing the SVM model. Specifically, they take a random sample of unlabeled data at each step, and rank them according to their informativeness w.r.t. current hyperplane. They disregard specific adaptations to class imbalance, by observing that imbalance is less severe in the class separating regions, under the assumption of locally Gaussian-distributed data classes.
In a classical uncertainty sampling scheme, imbalance is addressed using a boosting step \cite{zhu_active_2007}. However, the technique works under the assumption that a sufficiently large labeled data set is available initially, in order to train an acceptable initial model. As fit with imbalanced problems, they use recall as a performance metric.

Observing that cold start situations remain mostly unaddressed by the active learning literature, and are critical in the context of imbalanced problems, in  \cite{brangbour_active_2020} the general idea is to use a clustering structure to guide batch sampling. The underlying hypothesis is that the label information acquired in high quality clusters would be more effectively propagated to unlabeled elements using a semi-supervised algorithm.
A cluster quality index is combined to an impurity index, reflecting the entropy of per-cluster collected ground truth labels. Clusters are ranked according to the index, and sampling is performed in top clusters. Ranks are updated at each active learning iteration.

\section{Proposed Methodology} \label{sec:methodology}

We assume the distribution skew characteristic of class imbalance is expected, which is the case in many real-life contexts (e.g. biomedical applications, fraud detection, network intrusion \cite{he_learning_2009}). 
Combining a clustering structure to partial supervision has been explored in a similar way (\cite{brangbour_active_2020} and to some extent \cite{kang_using_2004}). However, with highly imbalanced and complex data sets (i.e. overlapping classes, multiple subconcepts), the likelihood that only labels from the majority class are returned for a sample of any cluster is high, which causes the proposed impurity criterion \cite{brangbour_active_2020} to discard clusters featuring more than average minority class content too early. Alternatively, using a semi-supervised model will allow to gracefully integrate label feedback.


In the next section, we motivate the usage of the conductance criterion \cite{almeida_is_2011} as a cluster quality score, that allows us to rank clusters according to the information feedback expected if sampling elements from them in the next active learning step. We then shortly review the soft-supervised model proposed by Subramanya and Bilmes \cite{subramanya_soft-supervised_2008}, that acts as the label propagation model updated at each active learning step in view to estimate entropies associated to unlabeled elements, and thus implement a per-element score used for sampling elements to send to the oracle.

\subsection{Cluster Quality Criterion} \label{sec:cluster}
Let us consider a partially labeled data set $\mathcal X = \{ x_i \}_{i \in \llbracket 1, N \rrbracket}$. We also consider $C$ possible ground truth classes for elements, indexed with integers from 1 to $C$. We build the set $\mathcal T = \{ t_i \}_{i \in \llbracket 1, N \rrbracket}$ of partial labels, with 0 as the placeholder for missing labels, and $t_i \in \llbracket 1, C\rrbracket$ otherwise. The set of labeled elements is then $\mathcal L = \{ x_i \}_{t_i > 0}$. In the cold start context considered in this paper, this set will be initially empty.

Let us also assume a clustering algorithm has been applied to the data set $\mathcal X$, yielding $K$ clusters, and a set of $N$ cluster labels $\mathcal Y = \{ y_i \}_{i \in \llbracket 1, N \rrbracket}$ with $y_i \in \llbracket 1, K \rrbracket$. 
Few unsupervised quality criteria can be decomposed per-cluster in a straightforward manner \cite{brangbour_active_2020}, thus enabling the computation of individual cluster quality scores. The conductance \cite{almeida_is_2011} offers this possibility, while having been recognized as showing good agreement to supervised criteria in the context of graph clustering \cite{emmons_analysis_2016}, and no explicit requirement of spherical clusters.
Using it requires converting pairwise Euclidean distances $d_{ij}$ between elements in $\mathcal X$ to edge weights (or similarities) $w_{ij}$ scaled in $[0,1]$, which can be easily performed e.g. using $w_{ij} = e^{-d_{ij}^2}$. 
Let $k \in \llbracket 1, K \rrbracket$ be a cluster index. Let also $\mathcal I_k$ the set of members of cluster $k$, i.e. the subset of $\mathcal X$ so that $x_i \in \mathcal I_k \text{ iif } y_i = k$. The conductance of cluster $k$ is defined as:

\vspace{-3mm}
\begin{equation}
    \Phi_k = 1 - \frac{\sum_{x_i \in \mathcal I_k, x_j \not\in \mathcal I_k} w_{ij}}{\min(a_k, a_{\bar k})}
\end{equation}

with $a_k = \sum_{x_i \in \mathcal I_k, x_j \in \llbracket 1, N \rrbracket} w_{ij}$ and $a_{\bar k} = \sum_{x_i \not\in \mathcal I_k, x_j \in \llbracket 1, N \rrbracket} w_{ij}$. Maximal conductance means a dense and well separated cluster. Per-cluster conductance already ranges in $[0,1]$, e.g. the overall conductance of a clustering is obtained by averaging per-cluster conductance. This means conductance can directly be used as a quality criterion to rank clusters. Also, conductance accounts for cluster size, i.e. intrinsically to its definition, small outlying but compact clusters will have low conductance. We verified this is indeed the case with validation experiments using UCI data sets \cite{Dua:2019}.

\subsection{Label Propagation Step} \label{sec:labelprop}

In this section we recall the semi-supervised label propagation model proposed by Subramanya and Bilmes \cite{subramanya_soft-supervised_2008}. It is closely related to \cite{zhu_semi-supervised_2003}, but in addition explicitly assigns class probability distributions to elements, on which our algorithms proposed in the next section rely. Label probability distributions $p_i$ and $q_i$ are hence defined for element $x_i \in \mathcal X$, so that their $c^\text{th}$ element $p_{i}(c) = q_{i}(c) = P(t_i = c)$, with $c \in \llbracket 1, C \rrbracket$. Also, probability distributions $r_i$ for elements in the labeled set $\mathcal L$ are defined so that $r_{i}(c) = 1 \text{ iif } t_i = c$, 0 else. For the consistency of definitions, $r_i$ is defined for all elements in $\mathcal X$, but is always 0 for unlabeled elements.
Then they define and minimize the following objective function:

\vspace{-5mm}
\begin{equation}
    \mathcal C(p,q) = \sum_{x_n \in \mathcal L} \text{KL}(r_i || q_i) + \mu \sum_{x_i \in \mathcal X}\sum_{x_j \in \mathcal X} w_{ij}' KL(p_i || q_j) - \nu \sum_{x_i \in \mathcal X} H(p_i) \label{eq:objective}
\end{equation}

with $KL(p||q)$ the Kullback-Leibler divergence of $p$ w.r.t. $q$, $H(p)$ the entropy of $p$, $w_{ij}' = w_{ij} + \alpha. \delta(i=j)$, and $\delta$ the function equalling 1 if its parametrized condition is verified, 0 else. Intuitively, the objective function pushes probability distributions of neighboring points to be as similar as possible, while enforcing the labels provided in $\mathcal L$ as much as possible. The last term ensures that maximal entropy (i.e. $p_{i}(c) = \nicefrac 1 C \text{ } \forall c$) is set by default when no neighboring information is available. We see that the algorithm depends on 3 hyper-parameters: $\mu$, $\nu$ and $\alpha$.
This objective function is convex, and can be estimated with the alternating minimization algorithm, that iterates the following closed form update formulas until convergence:

\vspace{-5mm}
\begingroup\makeatletter\def\f@size{8}\check@mathfonts
\def\maketag@@@#1{\hbox{\m@th\large\normalfont#1}}%
\begin{align}
    p^{(n)}_{i}(c) &\propto \exp \frac{\beta_i^{(n-1)}(c)}{\gamma_i} &
    q^{(n)}_{i}(c) &= \frac{r_i(c) + \mu \sum_{x_j \in \mathcal X} {w_{ji}'p_j^{(n)}(c)}}{\delta(x_i \in \mathcal L) + \mu \sum_{x_j \in \mathcal X} {w_{ji}'}} \label{eq:pestimate} \\
    \gamma_i &= \nu + \mu \sum_{x_j \in \mathcal X}{w_{ij}'} \nonumber &
    \beta_i^{(n-1)}(c) &= -\nu + \mu \sum_{x_j \in \mathcal X}{w_{ij}'(\log q_j^{(n-1)}(c) - 1)} \nonumber
\end{align}\endgroup

The $n$ superscript denotes estimates at iteration $n$. As means to be consistent with the entropy term in Equation \eqref{eq:objective}, all $p_i$ and $q_i$ are initialized with $\nicefrac 1 C \text{ } \forall c$.
Let us note that $p_i$ and $q_i$ (instead of only $p_i$) are mathematical artifacts meant to enable this algorithm. They prove that at convergence $p=q$, so both can be used afterwards.

In the present paper, rather than predicting classes for unlabeled elements, we are interested in selecting elements to be annotated by an oracle. The entropy terms reflects to which extent we are uncertain about the label of an element, so elements with high associated entropy are good candidates for selection. Also, we note the semi-supervised model in this section does not account for imbalance, so its usage for class prediction would be highly biased in favor of the majority class in such a context. While incorporating class frequency priors as in \cite{zhou_learning_2003} could be studied, we emphasize our focus on class entropy in an active learning context. Correction for imbalance is not necessary as the ranking of elements w.r.t. estimated entropy will focus on areas where classes are overlapping, regardless of imbalance.
In the next section, we discuss how the introduced cluster quality criterion and label propagation step are combined into active learning strategies.

\subsection{Active Learning Strategies} \label{sec:alstrategies}

\begin{algorithm}
\DontPrintSemicolon
\KwIn{$\mathcal X$, $\{ \mathcal I_k \}$, $n_\text{query}$, $n_\text{pc}$, $n_\text{iter}$, $\alpha$, $\nu$, $\mu$}
\KwOut{$n_\text{query} \times n_\text{iter}$ elements in $\mathcal L$}
Compute similarities $\{ w_{ij} \}$ \; \label{line:sim}
Cluster scores $\gets \{ \phi_k \}$ \; \label{line:phik}
Sort cluster scores in decreasing order \;
$q \gets \emptyset$ \;
\For{$k' \in \nicefrac {n_\text{query}}{n_\text{pc}}$ first clusters} {
    $q \gets q \cup \{$ sample $n_\text{pc}$ elements from $ \mathcal I_{k'}\}$\;
}
$\mathcal L \gets $ oracle feedback for $q$\;
Initialize semi-supervised model $\mathcal M$ with maximal $H(p_i)$\;
Update $\mathcal M$ with $\mathcal L$\;

\For{$t \in 1 \dots n_\text{iter}$} { \label{line:loop}
    Update cluster scores using $H(p_i)$\;
    Sort cluster scores in decreasing order \;
    $q \gets \emptyset$ \;
    \For{$k' \in \nicefrac {n_\text{query}}{n_\text{pc}}$ first clusters} {
        $q \gets q \cup \{$ sample $n_\text{pc}$ elements from $ \mathcal I_{k'}\}$\;
    }
    $\mathcal L \gets $ oracle feedback for $q$\;
    Update $\mathcal M$ with $\mathcal L$\; \label{line:modelupdate1}
}
\caption{Strategy with cluster ranking}
\label{algo:clusterranking}
\end{algorithm}

By contrast to other active learning models from the literature (e.g. \cite{ertekin_learning_2007}), we aim at strategies focused on increasing the chances of querying elements from the minority class. 
Our intuition is that when little or no label information is available at the beginning of the active learning procedure, sampling preferably from data space regions with higher density is likely to yield more informative feedback for a label propagation model.
A classification model is maintained through the execution of the strategy, but its main objective is to obtain a labeled set with less imbalance, but still representative of the initial unlabeled data distribution.
The algorithm will then tend to ignore regions where a given (typically the majority) label can be propagated easily, and focus on elements with high associated entropy. These come either from yet unexplored regions in the data space, or regions featuring overlapping classes. Updating an actual semi-supervised model, instead of computing statistics from the oracle feedback, allows to rank elements according to their specific estimated entropy, and avoids aggressive discards. This design can be related to uncertainty sampling \cite{lewis_sequential_1994}, as unlabeled elements with maximal associated entropy can be thought as close to implicit decision boundaries. 
We propose several ways to use models described in Sections \ref{sec:cluster} and \ref{sec:labelprop}, summarized by Algorithms \ref{algo:clusterranking} and \ref{algo:elementranking}.

In particular, we emphasize that updates of model $\mathcal M$ at each active learning step (lines \ref{line:modelupdate1} and \ref{line:modelupdate2} in Algorithms \ref{algo:clusterranking} and \ref{algo:elementranking}, respectively) use the current estimates for $p$ and $q$ distributions (see Equations \eqref{eq:pestimate}). At each step, $\mathcal M$ integrates the elements that have just been added to $\mathcal L$ (materialized by $r_i$ distributions in Equation \eqref{eq:objective}).

\begin{algorithm}
\DontPrintSemicolon
\KwIn{$\mathcal X$, $\{ \mathcal I_k \}$, $n_\text{query}$, $n_\text{pc}$, $n_\text{iter}$, $\alpha$, $\nu$, $\mu$}
\KwOut{$n_\text{query} \times n_\text{iter}$ elements in $\mathcal L$}
Same as Algorithm \ref{algo:clusterranking} up to line \ref{line:loop} \;
\For{$t \in 1 \dots n_\text{iter}$} {
    Update element scores for $\mathcal X \setminus \mathcal L$ using $H(p_i)$ \;
    $q \gets n_\text{query}$ first elements in $\mathcal X \setminus \mathcal L$ w.r.t. scores\;
    $\mathcal L \gets $ oracle feedback for $q$\;
    Update $\mathcal M$ with $\mathcal L$\; \label{line:modelupdate2}
}
\caption{Strategy with element ranking}
\label{algo:elementranking}
\end{algorithm}

In our experiments in the next section, we test 3 alternative methods for computing the similarities (Line \ref{line:sim} in Algorithm \ref{algo:clusterranking}) that are used as input to both conductance and label propagation steps:
\begin{itemize}
    \item The cluster (\emph{C}) similarity: $w_{ij} = 1$ if elements $x_i$ and $x_j$ are in the same $\mathcal I_k$,
    \item The nearest neighbor (\emph{NN}) similarity: $w_{ij} = 1$ if $x_j$ is among the $\kappa$ nearest neighbors of $x_i$ w.r.t. Euclidean distance $d$, 0 else,
    \item The Radial Basis Function (\emph{RBF}) similarity with local scaling: $w_{ij} = \frac {\exp {-d^2}}{\sigma_i \sigma_j}$. This local scaling was proposed by Zelnik-Manor and Perona \cite{zelnik-manor_self-tuning_2005}. We used the 2\% quantile of the distribution of distances to $x_i$ as $\sigma_i$ \cite{bruneau_observations_2016}.
\end{itemize}

Using \emph{C} will estimate the value of using cluster membership itself as a similarity function. \emph{RBF} with local scaling accounts for the potential variations of density in the data set.
By focusing on neighborhoods without explicit account of distances, we expect that using the \emph{NN} similarity in Equation \eqref{eq:objective} will mechanically put less attention on the majority class, hence effectively accounting for class imbalance.
Also, we test 4 alternative methods for computing cluster and element scores after the first algorithm step (which always uses $\phi_k$, Line \ref{line:phik} in Algorithm \ref{algo:clusterranking}):
\begin{itemize}
    \item Entropy only cluster (\emph{EOC}) score for cluster $k$ is $\frac 1 {| \mathcal I_k |} \sum_{x_i \in \mathcal I_k}{H(p_i)}$,
    \item Weighted entropy cluster (\emph{WEC}) score for cluster $k$ is $\phi_k . \textit{EOC}$,
    \item Entropy only element (\emph{EOE}) score for element $x_i$ is $H(p_i)$,
    \item Weighted entropy element (\emph{WEE}) score for element $x_i \in \mathcal I_k$ is $\phi_k.H(p_i)$
\end{itemize}

These score computation variants mainly differ by their usage of entropy information alone, or combined to cluster quality scores. If entropy information is used alone, then clusters are only used for the first active learning step, when stricly no labels are known yet.


\section{Experimental Section} \label{sec:experiments}

\subsection{Use Case Description}

The proposed experimental use case focuses as detecting content posted by people who have witnessed hurricane Harvey, that hit Texas and Louisiana in August 2017\footnote{\url{https://en.wikipedia.org/wiki/Hurricane_Harvey}}. The data we collected are all Tweets with location fields overlapping an area of approximately 40.000 $\text{km}^2$ around Houston, posted in between August the nineteenth and September the twenty-first of 2017. For the experiments in this section, we randomly extracted 10.000 elements among tweets featuring at least 3 words, and a location field reflecting a surface smaller than 350 $\text{km}^2$.

We want to classify the tweets in 3 categories: Positively Relevant (PR) if the tweet reveals that the user is currently in a flooded area, Negatively Relevant (NR), if the tweet reveals that the user is not currently in a flooded area, or Irrelevant (IR), if the tweet contains no information regarding the flood. We manually annotated a random subset of 1820 elements among which 1669 are irrelevant, 96 are negatively relevant and 55 positively relevant. The remaining 8120 elements remained unlabeled. The annotated sample has approximately 9\% relevant (i.e. PR+NR) elements. A significant imbalance is thus expected overall.

In order to estimate a label proxy of the remaining unlabeled set, we trained a SVM classifier with 80\% of the annotated dataset and tested it on the entire annotated dataset. We compute the F1 score by considering IR as negative class and PR+NR as positive class. Over 10 runs we obtained an average F1 of 0.74. Then we used the classifier to predict the labels of the unlabeled dataset. The resulting proxy label population is (IR: 9377, PR: 109, NR: 514), which results in 6\% relevant content. The disclosure of these proxy labels is used as the oracle in our experiments.

\subsection{Data Representation} \label{sec:representation}

We analyze Twitter posts from two perspectives: their textual content, and their spatio-temporal coordinates. Applying classification or clustering algorithms to textual content typically requires a preprocessing step, where the piece of text is \emph{embedded} in a high-dimensional numerical space, hence facilitating further calculations. 
For the experiments, we use a character-based language model, specially adapted to Twitter content by using hashtag prediction as proxy task \cite{dhingra_tweet2vec:_2016}. Specifically, we use the 200-dimensional output vectors of a Transformer network \cite{vaswani2017attention} trained as specified in \cite{dhingra_tweet2vec:_2016} as textual embedding, as we found it led to better performance and denser vectors.

A flood being a natural disaster tied to space and time, we also extracted a spatio-temporal representation of the tweet, which consists in the latitude, longitude and surface on the location field, and the number of hours between the tweet and the first tweet from the dataset.
We extracted clusters using the two representation spaces taken independently. Hence, according to definitions used in Section \ref{sec:cluster}, we obtain two cluster label vectors, $\mathcal Y_\text{text}$ and $\mathcal Y_\text{st}$ (where \emph{st} stands for \emph{spatio-temporal}). $\mathcal Y_\text{st}$ is extracted using Gaussian mixtures estimated using the EM algorithm \cite{bishop_pattern_2006}, with the number of cluster chosen heuristically as the number of days that separates the first tweet from the last in the dataset. $\mathcal Y_\text{text}$ is obtained using Affinity Propagation \cite{frey_clustering_2007} combined to local scaling \cite{zelnik-manor_self-tuning_2005} in order to cope with local variations of data density in the language model space. $\mathcal Y_\text{text}$ features 29 clusters, and $\mathcal Y_\text{st}$ 28 clusters.

\subsection{Combining Cluster Quality Scores} \label{sec:combining}

Combining heterogeneous representation spaces as described in the previous section can be related to subspace clustering \cite{vidal_subspace_2011}. However, in our case the aim is to use spatio-temporally homogeneous clusters, so we rather consider the Cartesian product of two clustering structures. Formally, we build $\mathcal Y_{\text{text} \times \text{st}}$ out of independent cluster label vectors. 787 clusters result from this product, with many singletons or very small clusters.
Scores for clusters in $\mathcal Y_{\text{text} \times \text{st}}$ average scores of respective clusters in $\mathcal Y_\text{text}$ and $\mathcal Y_\text{st}$ computed separately. 
The text representation being 200 dimensions and the spatio-temporal 4 dimensions, we rescale their associated distance matrices w.r.t. $\chi^2$ distributions \cite{bruneau_probabilistic_2018} in order to have comparable conductance scores.

\subsection{Results and Interpretation}

As our goal is to boost chances of labeling minority class elements with a limited annotation budget, we are evaluating our strategies by monitoring the recall of minority classes throughout the active learning process. At each iteration we keep the track of the acquired labels and update the metric. Let $\mathcal L_p$ and $\mathcal L_n$ be the set of PR and NR proxy labels, and $\mathcal L_t$ the set of all annotations returned by the oracle up to iteration $t$. For each iteration we can compute the recall scores of minority classes as $r_t = \frac{|\mathcal L_t \bigcap (\mathcal L_p \bigcup \mathcal L_n)|}{|(\mathcal L_p \bigcup \mathcal L_n)|}$.

\begin{figure}
  \caption{Recall curves for minority classes as a function of active learning iteration.}
  \label{global}
  \centering
  \includegraphics[width=0.8\columnwidth]{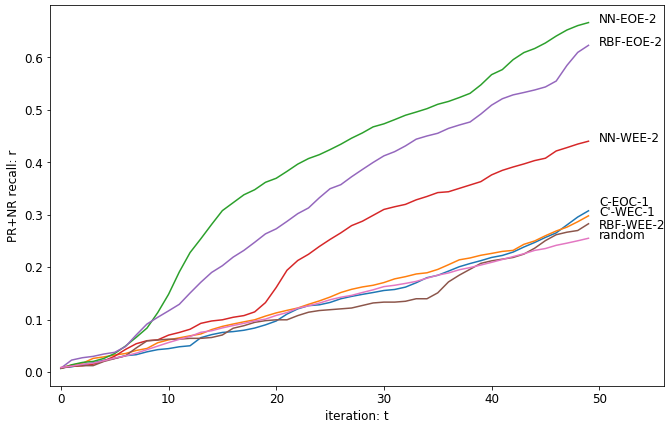}
\end{figure}

We ran Algorithms \ref{algo:clusterranking} and \ref{algo:elementranking} with the following parameters: $n_\text{query} = 50$, $n_\text{iter} = 50$, $n_\text{pc} = 3$. $\alpha$, $\nu$ and $\mu$ were set with 2, $10^{-3}$ and $10^{-3}$ as recommended by Subramanya and Bilmes \cite{subramanya_soft-supervised_2008}. In Figure \ref{global}, we compare 6 algorithmic variants named according to the nomenclature given in Section \ref{sec:alstrategies} (e.g. NN-EOE-2 designates Algorithm \ref{algo:elementranking} with NN similarity and EOE score function). As a reference, we also show the curves obtained if query elements are sampled completely at random. Each experiment proceeds until ${\nicefrac 1 4}^\text{th}$ of the data set is annotated by the oracle. The displayed curves average the results of 5 independent experiments.

Strategies using the cluster similarity (i.e. prefixed by \emph C) exhibit no significant performance improvement compared to randomness. Cluster membership hence does not appear to convey valuable information regarding the classification task at hand, and Algorithm \ref{algo:clusterranking}, based on sampling w.r.t. per-cluster ranks is not revealed as effective.
Algorithm \ref{algo:elementranking} with WEE scores (i.e. which weigh element-wise entropies with respective cluster conductances) also yields disappointing results. With the \emph{RBF} similarity, this algorithmic variant actually does not perform better than random. However, it becomes significantly better than random performance when used in conjunction with the \emph{NN} similarity, with almost twice more minority class elements retrieved by the end of the active learning iterations. 


The best performance is reached by Algorithm \ref{algo:elementranking} when using only entropies estimated by the semi-supervised model as scores for sample selection. Again, the \emph{NN} similarity significantly outperforms the \emph{RBF} similarity. In the end, \emph{NN-EOE-2} reaches more that 60\% recall of the minority classes with limited budget, which is approximately 3 times more than the random strategy. Overall, it appears that ranking w.r.t. entropies returned by the semi-supervised model already integrate the ability to effectively explore the data set. Further weighting by cluster quality scores yields significant degradation in the course of the algorithm, limiting its potential usefulness to the starting stage. We also see that \emph{NN}-based variants systematically overperform w.r.t. their \emph{RBF}-based counterpart, which confirms the adequacy of \emph{NN} similarity in an imbalanced context. 


\section{Conclusion}

A first important contribution in this paper has been to expose the peculiarities when trying to combine active learning methods to imbalanced cold-start classification problems. In this context, we designed active learning strategies, where we hypothesized a synergy between the exploitation of a clustering structure and a semi-supervised label propagation model. Several similarity functions were tested as means to effectively address class imbalance. We compared several algorithmic variants on a real-world imbalanced classification problem related to detecting information related to floods on Twitter.
In particular, we disclosed a way to combine the information carried by multiple subspaces, illustrated by our use case about extracting spatio-temporally aware information from a Twitter corpus. In the end, despite the absence of improvement brought by the clustering structure, an effective algorithmic variant was highlighted, which tripled the recall of minority class elements in a heavily imbalanced classification problem.

A possible reason for the ineffectiveness of using the clustering structure in our experiments may be its poor alignment with the minority class density. Ideally, we would like the granularity of the clustering structure guiding the active learning strategy to adapt to the label feedback. An idea would be to exploit the dendrogram resulting from a hierarchical clustering method, such as Hierarchical Agglomerative Clustering (HAC).

\section{Acknowledgements} \label{sec:ack}

This work was performed in the context of the Publimape project, funded by the CORE programme of the Luxembourgish National Research Fund (FNR).

\bibliographystyle{splncs04}
\bibliography{refs}

\end{document}